\documentclass[lettersize,journal]{IEEEtran}
\usepackage{amsmath,amsfonts}
\usepackage{algorithmic}
\usepackage{algorithm}
\usepackage{array}
\usepackage[caption=false,font=footnotesize,labelfont=rm,textfont=rm]{subfig}
\usepackage{textcomp}
\usepackage{stfloats}
\usepackage{url}
\usepackage{verbatim}
\usepackage{graphicx}
\usepackage{bbding}
\usepackage[table]{xcolor}
\usepackage{tablefootnote}
\usepackage{cite}

\usepackage{hyperref}
\hypersetup{hypertex=true,
	colorlinks=true,
	linkcolor=blue,
	anchorcolor=blue,
	citecolor=blue}
\usepackage{booktabs}
\usepackage{makecell}
\usepackage{booktabs}
\usepackage{bbm}
\usepackage{multirow}

\usepackage{pifont}

\usepackage[capitalize]{cleveref}
\makeatother

\begin{document}
\captionsetup{font={small}}	

\title{FMRT: Learning Accurate Feature Matching with Reconciliatory Transformer}

\author{Xinyu Zhang$^*$, Li Wang$^*$, Zhiqiang Jiang, Kun Dai, Tao Xie, Lei Yang, Wenhao Yu, Yang Shen, Jun Li

\thanks{This work was supported by the National High Technology Research and Development Program of China under Grant No. 2018YFE0204300, and the National Natural Science Foundation of China under Grant No. 62273198, U1964203, 52221005. \emph{(Corresponding author: Li Wang. *: These authors contributed equally to this work.)}}

\thanks{Xinyu Zhang, Li Wang, Lei Yang, Wenhao Yu, Yang Shen and Jun Li are with the School of Vehicle and Mobility, Tsinghua University, Beijing 100084, China (e-mail: 
xyzhang@tsinghua.edu.cn;
lwang\_hit@hotmail.com;
yanglei20@mails.tsinghua.edu.cn;
wenhaoyu@mail.tsinghua.edu.cn;
shenyang 5008@163.com;
lijun19580326@126.com).
}

\thanks{Zhiqiang Jiang, Kun Dai, and Tao Xie are with State Key Laboratory of Robotics and System, Harbin Institute of Technology, Harbin 150006, China (e-mail: jzq@stu.hit.edu.cn; 20s108237@stu.hit.edu.cn; xietao1997@hit.edu.cn).
}
}

	
\maketitle

\begin{abstract}
Local Feature Matching, an essential component of several computer vision tasks (e.g., structure from motion and visual localization), has been effectively settled by Transformer-based methods. However, these methods only integrate long-range context information among keypoints with a fixed receptive field, which constrains the network from reconciling the importance of features with different receptive fields to realize complete image perception, hence limiting the matching accuracy. In addition, these methods utilize a conventional handcrafted encoding approach to integrate the positional information of keypoints into the visual descriptors, which limits the capability of the network to extract reliable positional encoding message. In this study, we propose Feature Matching with Reconciliatory Transformer (FMRT), a novel Transformer-based detector-free method that reconciles different features with multiple receptive fields adaptively and utilizes parallel networks to realize reliable positional encoding. Specifically, FMRT proposes a dedicated Reconciliatory Transformer (RecFormer) that consists of a Global Perception Attention Layer (GPAL) to extract visual descriptors with different receptive fields and integrate global context information under various scales, Perception Weight Layer (PWL) to measure the importance of various receptive fields adaptively, and Local Perception Feed-forward Network (LPFFN) to extract deep aggregated multi-scale local feature representation. 
Besides, we introduce a novel Axis-Wise Position Encoder (AWPE) that perceives positional encoding as two independent keypoints encoding tasks along the row and column dimensions and utilizes two parallel network branches to decouple the $X$- and $Y$-coordinate information of keypoints, so that boosting the capability of the network to model position information for different images.
Extensive experiments demonstrate that FMRT yields extraordinary performance on multiple benchmarks, including pose estimation, visual localization, homography estimation, and image matching.


\end{abstract}

\begin{IEEEkeywords}
Local feature matching, transformer, match refinement.
\end{IEEEkeywords}

\section{Introduction}
\IEEEPARstart{L}{ocal} feature matching is a crucial part of several robot applications, such as Structure from Motion (SFM) \cite{schonberger2016structure, mousavi2022two, li2022pulse}, Simultaneous Localization and Mapping (SLAM) \cite{campos2021orb, li2023whu, wan2022terrain, junior2022ekf, zhao2022study, pairet2022online}, and visual localization \cite{sun2023f3, taira2018inloc, sarlin2019coarse}.

As a widely investigated pipeline, the detector-based methods \cite{lowe2004distinctive, rublee2011orb, DeTone_2018_CVPR_Workshops, Dusmanu_2019_CVPR, revaud2019r2d2, tyszkiewicz2020disk, sarlin2020superglue, xia2022locality, chen2021learning, kuang2021densegap, shi2022clustergnn} leverage the elaborate detectors to detect keypoints, utilize handcrafted \cite{rublee2011orb, lowe2004distinctive} or learning-based vectors \cite{detone2018superpoint, DeTone_2018_CVPR_Workshops, revaud2019r2d2, luo2020aslfeat, zhao2022alike} to describe them, and employ a matching algorithm (e.g., mutual nearest neighbor) \cite{bian2017gms, zhang2019learning, tao2002continuous, chen2022csr} to extract correspondences.
Although realizing impressive matching results, this pipeline yields inferior performance in several environments with substantial indistinctive regions since the detectors cannot capture repeatable keypoints in image pairs.

Current with the detector-based methods, detector-free methods \cite{rocco2018neighbourhood, rocco2020efficient, li2020dual, zhou2021patch2pix, efe2021dfm, sun2021loftr, wang2022matchformer, giang2023topicfm, chen2022aspanformer, shen2022semi} realize cutting-edge matching performance in severe environment.
The early approaches \cite{rocco2018neighbourhood, rocco2020efficient, li2020dual, zhou2021patch2pix, efe2021dfm} utilize convolutional neural networks (CNNs) to extract keypoints with visual descriptors across dense grids of images.
However, due to the limited receptive field, these methods have difficult in distinguishing keypoints with similar feature representations.
Recently, Transformer \cite{vaswani2017attention} has achieved excellent performance in several computer vision tasks due to its strong capability to model long-range dependencies, which inspires the researchers to leverage Transformer to boost matching accuracy \cite{sun2021loftr, wang2022matchformer, giang2023topicfm, chen2022aspanformer}.
As a representative work, LoFTR \cite{sun2021loftr} leverages the self and cross-attention mechanism in Linear Transformer \cite{katharopoulos2020transformers} to enhance the visual descriptors of keypoints with manageable computational cost.
Although the Transformer-based detect-free methods exhibit excellent capability, two issues warrant further exploration.

\begin{figure}[]
    \centering
    \includegraphics[width=1.0\hsize]{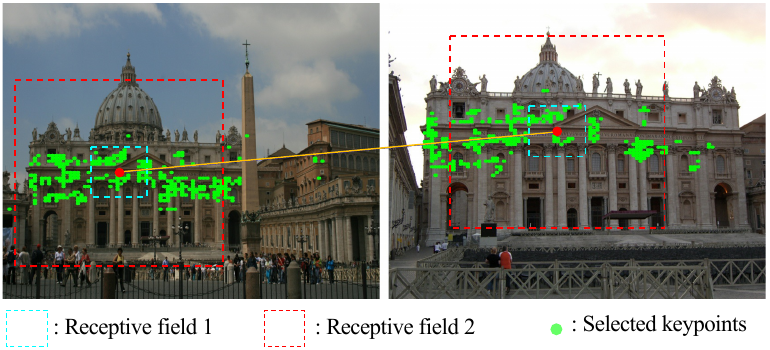}
    \caption{Visualization of two receptive fields. We visualize 300 keypoints (green points) that provide the most predominant information when using a Transformer to integrate long-range context.
    By comparing the two receptive fields, we observe that receptive field 1 contains dense but inadequate keypoint information, while receptive field 2 contains sufficient but sparse keypoint information.
    So it is meaningful to employ various receptive fields and reconcile them well.
   } 
    \label{pipeline}
\end{figure}

(i) As illustrated in \cref{pipeline}, we visualize the keypoints that provide the most abundant information when leveraging Transformer to integrate global context.
It can be seen that the features with various receptive fields involve different amounts of crucial information.
However, existing methods only utilize CNN to extract keypoints with a fixed receptive field and integrate long-range context information among them, which leads to inferior image perception since the plentiful essential message is discarded or not emphasized.
Besides, diverse images exhibit distinct responses to the features with different receptive fields.
Therefore, it is essential to investigate an approach that enables the network to extract the visual descriptors with various receptive fields and reconcile them adaptively.

(ii) In current methods (e.g., LoFTR \cite{sun2021loftr}, MatchFormer \cite{wang2022matchformer}), absolute sinusoidal positional encoding is utilized to incorporate the positional information into the visual descriptors of keypoints.
However, the handcrafted positional information may not be ideal for local feature matching.
Recently, neural networks have exhibited remarkable capability to model image information.
Hence, it is significant to leverage the neural network to extract the optimal positional encoding information.


To solve the above issues, we propose a novel Transformer-based detector-free method named Feature Matching with Reconciliatory Transformer (FMRT) that captures dense and precise correspondences.
For the first issue, FMRT proposes \textbf{Reconciliatory Transformer (RecFormer)} that adaptively reconciles features with different receptive fields to generate discriminative features.
RecFormer is comprised of Global Perception Attention Layer (GPAL), Perception Weight Layer (PWL), and Local Perception Feed-forward Network (LPFFN).
As the first part, GPAL utilizes parallel depth-wise convolution \cite{howard2017mobilenets} to extract visual descriptors with different receptive fields, followed by two parallel Linear Attention Layer \cite{katharopoulos2020transformers} to aggregate global context information across keypoints under different scales.
After that, PWL utilizes MLP and softmax to generate weight coefficients, which are leveraged to adaptively reconcile the importance of different features for the local feature matching task.
Ultimately, considering Transformer only models long-range global dependencies among keypoints, LPFFN stacks the parallel depth-wise convolutions with different kernel sizes sequentially to extract multi-scale local features progressively. 
By interleaving the RecFormer to implement self- and cross-attention multiple times, FMRT comprehensively reconciles the impact of features with different receptive fields, hence deriving the optimal features for the subsequent correspondence prediction task.

For the second issue, FMRT puts forward a novel \textbf{Axis-Wise Position Encoder (AWPE)} that utilizes parallel networks to capture superior positional encoding information for local feature matching tasks.
We perceive the positional encoding as two independent keypoints encoding tasks along the row and column dimensions and utilize two sets of Conv1d layers to project the 1D X- and Y-coordinates of keypoints into two independent high-dimension vectors.
Then, AWPE adds the X- and Y-features together to derive positional encoding maps, which are integrated into the coarse-level features to model the position information.
As shown in \cref{ablation_positional}, compared with other positional encoding methods, AWPE boosts the local feature matching capability of the network.

We summarize the contributions of this work as follows:

\begin{itemize}
    \item{We propose a Reconciliatory Transformer that extracts the features with diverse receptive fields and reconciles them adaptively to generate discriminative visual descriptors, hence elevating the modeling capability of the network.}
    \item{We develop an Axis-Wise Position Encoder that utilizes a parallel architecture to boost the capability of the network to model positional encoding information.}
    \item{We demonstrate that FMRT achieves extraordinary performance on all homography estimation, image matching, relative pose estimation, and visual localization tasks.}
\end{itemize}

\section{Related Work}
\subsection{Detector-based Local Feature Matching}
The conventional detector-based methods \cite{lowe2004distinctive, rublee2011orb, DeTone_2018_CVPR_Workshops, Dusmanu_2019_CVPR, revaud2019r2d2, tyszkiewicz2020disk, sarlin2020superglue, xia2022locality, chen2021learning, kuang2021densegap, shi2022clustergnn} follows the pipeline that detects, describes, and matches keypoints.
Being long investigated, these methods have realized a decent balance between inference speed and matching accuracy, and become the dominant approach for local feature matching.

The traditional hand-crafted feature matching algorithms (e.g., SIFT \cite{lowe2004distinctive} and ORB \cite{rublee2011orb}) are integrated into several mature robotics algorithms (e.g., ORB-SLAM series \cite{mur2017orb, campos2021orb}).
However, owing to the inferior handcrafted visual descriptors, these methods suffer from severe performance degradation in harsh environments with low textures and dramatic changes in light intensity.
With the development of deep learning, many approaches \cite{DeTone_2018_CVPR_Workshops, Dusmanu_2019_CVPR, revaud2019r2d2, tyszkiewicz2020disk, zhao2022alike, luo2020aslfeat} elaborately design CNNs to extract robust and descriminative visual descriptors, hence significantly improving the matching performance.
SuperPoint \cite{detone2018superpoint} utilizes a self-supervised domain adaptation framework for keypoints detection and description. 
D2Net \cite{Dusmanu_2019_CVPR} postpones the keypoints detection stage until reliable information is available.
Nevertheless, these methods have difficulty in differentiating similar keypoints due to the limited receptive field of CNNs.


Recently, Transformer \cite{vaswani2017attention} has achieved excellent performance in various computer vision tasks owing to its powerful capability to integrate global context information \cite{wang2022hybrid, dai2022ao2, li2022sdtp, chen2022transformer, liu2021swinnet, zhu2022mlst,yao2023dual, wu2023meta}.
Inspired by this, SuperGlue \cite{sarlin2020superglue} leverages SuperPoint to extract keypoints with visual descriptors, and utilizes the attention mechanism to aggregate global information intra- / inter-images.
However, the computational consumption of the vanilla Transformer \cite{liu2021swin} makes SuperGlue occupy massive computing resources when processing a large number of keypoints.
To address this problem, plentiful approaches \cite{chen2021learning, kuang2021densegap, shi2022clustergnn, suwanwimolkul2022efficient, liu2022fgcnet} investigate to optimize the Transformer architecture of SuperGlue.
SGMNet \cite{chen2021learning} captures a small number of seeds as the message passing bottleneck to reduce the computational cost of Transformer.
Although exhibiting excellent matching performance, these detector-based methods cannot establish satisfactory correspondences even with perfect matching algorithms if the detectors fail to extract repeatable keypoints.


\begin{figure*}[]
    \centering
    \includegraphics[width=0.999\hsize]{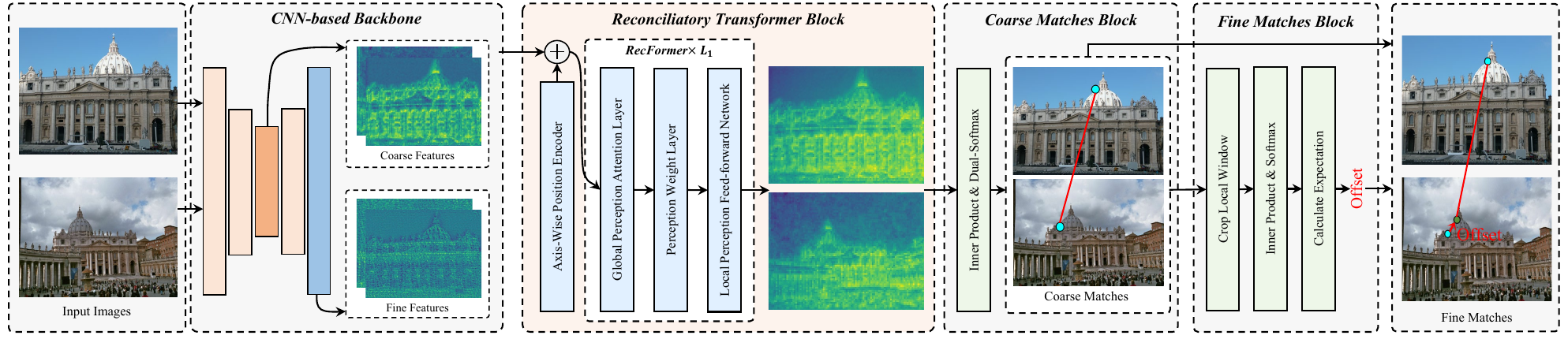}
    \caption{Overview of the proposed FMRT. FMRT first utilizes a \textbf{CNN-based Backbone} to concurrently extract coarse and fine features. Then, \textbf{Reconciliatory Transformer Block} is proposed to reconcile the importance of features with different receptive fields, hence effectively elevating the modeling capability of the network.
    Ultimately, \textbf{Coarse Matches Block} and \textbf{Fine Matches Block} are proposed to generate final correspondences in a coarse-to-fine manner.} 
    \label{overall}
\end{figure*}
\subsection{Detector-free Local Feature Matching}
Different from the detector-based methods, the detector-free methods \cite{rocco2018neighbourhood, rocco2020efficient, li2020dual, zhou2021patch2pix, efe2021dfm, sun2021loftr, wang2022matchformer, giang2023topicfm, chen2022aspanformer, xie2023deepmatcher, dai2023oamatcher} discards the feature extraction stage and directly generate matches from the dense pixels of images.
Specifically, the detector-free methods establish the matches among pixels, rather than the extracted keypoints.
Therefore, massive repeatable keypoints are extracted in the overlapping areas of image pairs, hence elevating the matching performance noticeably.

Earlier detector-free works \cite{rocco2020efficient, li2020dual, zhou2021patch2pix, efe2021dfm} utilize cost volume to enumerate all candidate correspondences.
DRC-Net \cite{li2020dual} calculates a 4D cost volume to generate coarse matches, which are refined by a learnable neighborhood consensus module.
Patch2Pix \cite{zhou2021patch2pix} utilizes CNNs to detect coarse matches in low-resolution feature maps, and then optimize them at higher resolution.
DFM \cite{efe2021dfm} uses a VGG architecture \cite{simonyan2014very} to extract features and improve matching performance without requiring additional training.
Although elevating the matching performance, the limited respective fields of CNNs make these methods have difficult to discriminate incorrect correspondences with similar structure information.

Recently, after witnessing the success of SuperGlue, the cutting-edge detector-free methods \cite{sun2021loftr, wang2022matchformer, giang2023topicfm, chen2022aspanformer, tang2022quadtree, xie2023deepmatcher, dai2023oamatcher} focuses on realizing global consensus with the help of Transformer.
As a representative work, LoFTR \cite{sun2021loftr} utilizes Linear Transformer to model long-range global dependencies and update features of all keypoints, thereby achieving outstanding performance and ensuring manageable computation costs.
After that, substantial works follow the design of LoFTR and make several improvements.
MatchFormer \cite{wang2022matchformer} utilizes the attention at each stage of the encoder to conduct feature extraction and feature matching concurrently.
QuadTree \cite{tang2022quadtree} introduces a novel Transformer architecture that calculates hierarchical attention by constructing a token pyramid.
ASpanFormer \cite{chen2022aspanformer} integrates the optical flow estimation into local feature matching task, hence making the network perform feature interaction on the local regions.
DeepMatcher \cite{xie2023deepmatcher} designs a deep Transformer architecture to extract easy-to-match features.
OAMatcher \cite{dai2023oamatcher} leverages overalpping areas to realize efficient image propagation.
However, existing methods fix the visual descriptors to a single receptive field, making them discard substantial crucial information.
Besides, the absolute sinusoidal positional encoding used in current methods provides weak position modeling information, which greatly limits the capability of the network to handle local feature matching tasks.
In this work, we propose FMRT that enables the method adaptively reconciles the importance of features with various receptive fields, and extracts abundant position modeling information for different images.

\subsection{Efficient Transformer}
Recently, Transformer has become popular in the realm of computer vision tasks, such as image classification, and object detection.
Vanilla Transformer \cite{dosovitskiy2020image} leverages an attention matrix to integrate global context information, which makes it difficult to handle long sequences since the computation complexity is quadratic to the length of sequences.
After that, plentiful methods have been proposed to construct efficient Transformer architecture.
Linear Transformer \cite{katharopoulos2020transformers} utilizes a nonlinear function to fit the softmax operation and leverages the associativity property to control the computational consumption.
FastFormer \cite{wu2021fastformer} utilizes an additive attention mechanism that utilizes a global attention vector to realize linear complexity.
DCT-Former \cite{scribano2023dct} utilizes the properties of the Discrete Cosine Transform to approximate the attention module.

\section{Methodology}

\subsection{Overview}
As illustrated in \cref{overall}, We present the overall structure of FMRT.
Taking an image pair $I_{A}, I_{B}$ as input, FMRT designs four steps to establish accurate matches.

(i) \textbf{CNN-based Backbone} is proposed to concurrently extract coarse features $\bar{F}_{A}, \bar{F}_{B}$ and fine features $\tilde{F}_{A}, \tilde{F}_{B}$.
(ii) \textbf{Reconciliatory Transformer Block (RTB)} first utilizes \textbf{Axis-Wise Position Encoder (AWPE)} to aggregate decoupled positional information into coarse features, obtaining the discriminative visual descriptors.
Then, RTB converts the 2D features to 1D sequences and interleaves the Reconciliatory Transformer (RecFormer) to model global dependencies among all keypoints.
(iii) \textbf{Coarse Matches Block} and \textbf{Fine Matches Block} are introduced to extract final matches in a coarse-to-fine manner.

 
\subsection{CNN-based Backbone}
As the first part, we employ ResNet \cite{he2016deep} with FPN \cite{lin2017feature} to extract initial coarse features $\bar{F}_A$, $\bar{F}_B \in \mathbb{R}^{\bar{C} \times H/8 \times W/8}$ and fine features $\tilde{F}_A$, $\tilde{F}_B \in \mathbb{R}^{\tilde{C} \times H/2 \times W/2}$, with $H$ and $W$ being the height and width of the images.
Following other detector-free methods, the input images are divided by $8 \times 8$ grids, whose central pixels are viewed as keypoints $P_{A}, P_{B} \in \mathbb{R}^{N \times 2}$, where $N = H/8 \times W/8$.


\begin{figure*}[]
    \centering
    \includegraphics[width=0.99\hsize]{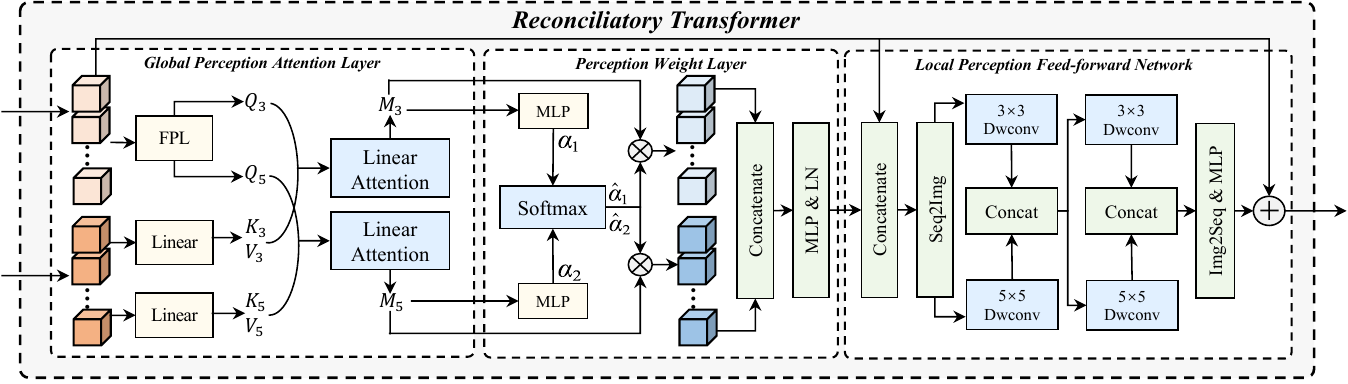}
    \caption{The overall architecture of Reconciliatory Transformer (RecFormer). RecFormer is comprised of Global Perception Attention Layer (GPAL), Perception Weight Layer (PWL), and Local Perception Feed-forward Network (LPFFN).} 
    \label{RecFormer}
\end{figure*}

\subsection{Reconciliatory Transformer Block (RTB)}
As shown in \cref{overall}, RTB consists of Axis-Wise Position Encoder (AWPE) and Reconciliatory Transformer (RecFormer).

\begin{figure}[]
    \centering
    \includegraphics[width=0.99\hsize]{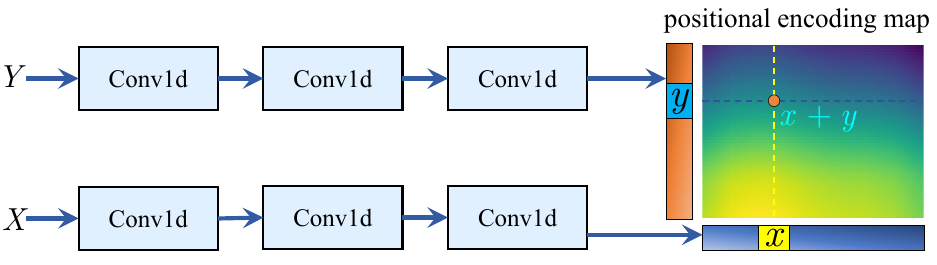}
    \caption{\textbf{The illustration of AWPE}. The $X$ and $Y$ vectors are encoded independently to generate a positional encoding map.} 
    \label{AWPE}
\end{figure}

\textbf{Axis-Wise Position Encoder (AWPE).}
Current detector-free approaches \cite{sun2021loftr, giang2023topicfm} utilize absolute sinusoidal positional encoding to incorporate the positional information into the visual descriptors of keypoints.
However, the handcrafted positional information may not be ideal for local feature matching.
Recently, the neural network has exhibited remarkable capability to model image information.
Hence, it is significant to leverage the neural network to extract the optimal positional encoding information.
Therefore, we propose an Axis-Wise Position Encoder (AWPE) that perceives the positional encoding as two independent keypoints encoding tasks along the row and column dimensions and utilize two independent sets of Conv1d layers to project the 1D X- and Y-coordinates into two independent high-dimension vectors, hence modeling the positional information.


As shown in \cref{AWPE}, AWPE utilizes two sets of Conv1d layers to map $X=[0,1,...,W/8-1] \in \mathbb{R}^{1 \times W/8}$ and $Y=[0,1,...,H/8-1] \in \mathbb{R}^{1 \times H/8}$ into two independent high-dimension vectors $F^{X} \in \mathbb{R}^{\bar{C} \times W/8}$ and $F^{Y} \in \mathbb{R}^{\bar{C} \times H/8}$.
After that, we extend the dimension of the $F^{X}$ and $F^{Y}$ to $R^{\bar{C} \times 1 \times W/8}, R^{\bar{C} \times H/8 \times 1}$ and add them together to derive positional encoding maps $F^{pos}_{A} \in \mathbb{R}^{\bar{C} \times H/8 \times W/8}$ and $F^{pos}_{B} \in \mathbb{R}^{\bar{C} \times H/8 \times W/8}$.
Finally, we add $F^{pos}_{A}, F^{pos}_{B}$ to coarse features $\bar{F}_A$, $\bar{F}_B$.
This process can be formulated as:
\begin{equation}
    \begin{aligned}
        \bar{F}_{A}^{p} = \bar{F}_A+F^{pos}_{A},\\
        \bar{F}_{B}^{p} = \bar{F}_B+F^{pos}_{B}.
    \end{aligned}
\end{equation}

Ultimately, we convert the 2D features $\bar{F}_A^{p}$, $\bar{F}_B^{p}$ to 1D sequence, deriving $\bar{F}_{A}^{seq}$, $\bar{F}_{B}^{seq} \in \mathbb{R}^{N \times \bar{C}}$.
Notably, we perceive $\bar{F}_{A}^{seq}$, $\bar{F}_{B}^{seq}$ as the visual descriptors of keypoints $P_{A}, P_{B}$.

\textbf{Reconciliatory Transformer (RecFormer).}
The diverse features with various receptive fields involve different amounts of information, and integrating these features into a discriminative visual descriptor is crucial to generate accurate matches.
However, existing methods \cite{sun2021loftr, wang2022matchformer, chen2022aspanformer} only utilize the Transformer to model the long-range dependencies among the keypoints with a fixed receptive field, making them discard the substantial essential message.
Besides, diverse images exhibit distinct responses to the features with different receptive fields.
To solve the aforementioned problems, we propose RecFormer which predicts weight coefficients to reconcile the importance of features with different receptive fields.
As shown in \cref{RecFormer}, RecFormer is comprised of Global Perception Attention Layer (GPAL), Perception Weight Layer (PWL), and Local Perception Feed-forward Network (LPFFN).



\begin{figure}[]
    \centering
    \includegraphics[width=0.99\hsize]{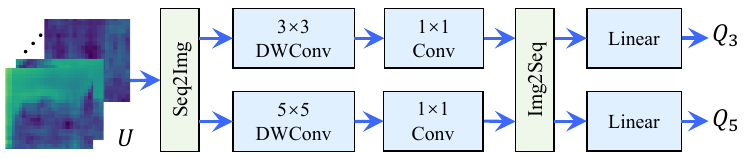}
    \caption{\textbf{The structure of FPL}. FPL takes features $U$ as input and predicts $Q_{3}$, $Q_{5}$.} 
    \label{FPL}
\end{figure}

\emph{Global Perception Attention Layer (GPAL).}
GPAL first utilizes Feature Perception Layer (FPL) to handle input features $U$ to extract features with different receptive fields.
Specifically, As shown in \cref{FPL}, FPL converts $U$ to 2D feature maps and employs two parallel $(3\times3, 5\times5)$ depth-wise convolutions to extract features with different receptive fields.
Then, FPL utilizes two parallel $1 \times 1$ point-wise convolution to squeeze the channel dimension and reshapes the results to 1D sequences $\breve{U}_{3}$, $\breve{U}_{5} \in \mathbb{R}^{N \times \bar{C}/2}$
Then, the features $\breve{U}$ and $R$ are projected into different query vectors $Q_{3}$, $Q_{5}$, key vectors $K_{3}$, $K_{5}$, value vectors $V_{3}$, $V_{5}$ by MLPs.
This process can be formulated as:
\begin{equation}
\begin{aligned}
Q_{3} &= MLP(I2S(C_{1}(DW_{3}(S2I(U))))), \\
Q_{5} &= MLP(S2I(C_{1}(DW_{5}(S2I(U))))), \\
&K_{3} = MLP(R), \ \ \ K_{5} = MLP(R), \\
&V_{3} = MLP(R), \ \ \ V_{5} = MLP(R), \\
\end{aligned}
\end{equation}
where $I2S(\cdot)$, $S2I(\cdot)$ mean converting features / sequences  to sequences / features; $C_{1}(\cdot)$ means $1 \times 1$ point-wise convolution.

Subsequently, following LoFTR, we utilize linear attention \cite{katharopoulos2020transformers} to model the long-range context dependencies among features with different receptive fields, which can be formulated as:
\begin{equation}
\begin{aligned}
    M_3 &= \phi(Q_3)(\phi(K_3)^{T}V_3),\\
    M_5 &= \phi(Q_5)(\phi(K_5)^{T}V_5),
\end{aligned}
\end{equation}
where $\phi(\cdot) = elu(\cdot) + 1$.

\emph{Perception Weight Layer (PWL).}
$M_3$, $M_5 \in \mathbb{R}^{N \times \bar{C}/2}$ represent the propagated message among keypoints under different receptive fields.
Considering various receptive fields involve different amounts of information, we propose PWL to predict weight coefficients, which are utilized to reconcile the importance of $M_3$, $M_5$ adaptively.
More concretely, MLPs are utilized to process $M_3$, $M_5$ to calculate the response values $\alpha_{1}$ and $\alpha_{2}$, followed by a softmax algorithm to generate weight coefficients $\hat{\alpha}_{1}$ and $\hat{\alpha}_{2}$:
\begin{equation}
\begin{aligned}
    (\hat{\alpha}_{1}, \hat{\alpha}_{2}) = Softmax(MLP(M_3), MLP(M_5)).
\end{aligned}
\end{equation}

Subsequently, PWL utilizes $\hat{\alpha}_{1}$ and $\hat{\alpha}_{2}$ to weight $M_3$, $M_5$, and then concatenates the results along the channel dimension.
Therefore, we ensure the network can automatically weigh the importance of different receptive fields.
Ultimately, we derive the reconciliatory message $\breve{M} \in \mathbb{R}^{N \times \bar{C}}$.
\begin{equation}
\begin{aligned}
    \breve{M} = LN(MLP([\hat{\alpha}_{1} M_3 || \hat{\alpha}_{2} M_5])),
\end{aligned}
\end{equation}
where $[\cdot||\cdot]$ means concatenating the features along channel dimension; $LN(\cdot)$ means layer normalization.


\emph{Local Perception Feed-forward Network (LPFFN).}
Considering Transformer only models global dependencies, Local Perception Feed-forward Network (LPFFN) stacks the parallel depth-wise convolutions sequentially to extract multi-scale local features progressively. 
Concretely, LPFFN first concatenates $\breve{M}$ and the initial input features $U$ along the channel dimension and converts the results to 2D feature maps $\breve{M_{l}} \in \mathbb{R}^{2 \bar{C} \times H/8 \times W/8}$:
\begin{equation}
\begin{aligned}
    \breve{M_{l}} = S2I([U||\breve{M}]).
\end{aligned}
\end{equation}

Then, LPFFN employs $(3\times3$, $5\times5)$ parallel depth-wise convolutions \cite{howard2017mobilenets} to extract multi-scale local feature presentations and concatenates them to generate intermediate feature representation $\breve{M_{l1}} \in \mathbb{R}^{4\bar{C} \times H/8 \times W/8}$
\begin{equation}
\begin{aligned}
    \breve{M_{l1}} = [ DW_{3}(\breve{M_{l}}) || DW_{5}(\breve{M_{l}}) ].
\end{aligned}
\end{equation}

Subsequently, LPFFN performs the above procedures again, converts sequences to feature maps, and utilizes an MLP to squeeze the channel dimension.
Ultimately, we design a shortcut structure to integrate the input feature $U$ into the final discriminative feature representation $M_{F} \in \mathbb{R}^{N \times \bar{C}}$.
\begin{equation}
\begin{aligned}
    M_{F} = U + MLP(I2S([ DW_{3}(\breve{M_{l1}}) || DW_{5}(\breve{M_{l1}}) ])).
\end{aligned}
\end{equation}

For convenience, we define the above procedures of RecFormer as:
\begin{equation}
    M_{F} = RecF(U, R).
\end{equation}

We utilize RecFormer to enhance the visual descriptors $\bar{F}_{A}^{seq}$, $\bar{F}_{B}^{seq}$ by $L_1$ times.
During the $l$-th feature enhancement, the input features $(U, R)$ are the same (either $(\bar{F}_{A}^{seq}, \bar{F}_{A}^{seq})$ or $(\bar{F}_{B}^{seq}, \bar{F}_{B}^{seq})$) for self-attention, and the different (either $(\bar{F}_{A}^{seq}, \bar{F}_{B}^{seq})$ or $(\bar{F}_{B}^{seq}, \bar{F}_{A}^{seq})$) for cross-attention.
This process can be formulated as:
\begin{equation}
    \begin{aligned}
        ^{l-1}\bar{F}^{seq}_{A} &= RecF(^{l-1}\bar{F}^{seq}_{A}, ^{l-1}\bar{F}^{seq}_{A}), \\
        ^{l-1}\bar{F}^{seq}_{B} &= RecF(^{l-1}\bar{F}^{seq}_{B}, ^{l-1}\bar{F}^{seq}_{B}), \\
        ^{l}\bar{F}^{seq}_{A} &= RecF(^{l-1}\bar{F}^{seq}_{A}, ^{l-1}\bar{F}^{seq}_{B}), \\
        ^{l}\bar{F}^{seq}_{B} &=RecF(^{l-1}\bar{F}^{seq}_{B}, ^{l}\bar{F}^{seq}_{A}).
    \end{aligned}
    \label{4steps}
\end{equation}

Ultimately, we obtain the final visual descriptors $^{L}\bar{F}_{A}^{seq}$ and $^{L}\bar{F}_{B}^{seq}$. 


\subsection{Coarse Matches Block (CMB)}
Given $^{L}\bar{F}_{A}^{seq}$ and $^{L}\bar{F}_{B}^{seq}$, we utilize the inner product to obtain score matrix $S \in \mathbb{R}^{N \times N}$, which are processed by dual-softmax \cite{sun2021loftr} to derive confidence matrix $G$.
\begin{equation}
    \begin{aligned}
        S(i,j) &= \langle ^{L}\bar{F}_{A}^{seq}, \ ^{L}\bar{F}_{B}^{seq}  \rangle, \\
        G(i,j) = Soft&max(S(i,\cdot))_{j} \cdot Softmax(S(\cdot,j))_{i}
    \end{aligned}
\end{equation}

where $\langle \cdot , \cdot \rangle$ means the inner product.

Subsequently, we select the matches that satisfy the mutual nearest neighbor (MNN) criteria and possess high confidence as the coarse matches.
According to the coordinates $P_{A}, P_{B}$ of keypoints, we define the coarse matches $E^{c} = \{(P_{A}^{c}, P_{B}^{c})\}$ as:
\begin{equation}
\begin{aligned}
    D = \{ &(i,j) | G(i,j) > \rho, \ \forall (i,j) \in MNN(G)\}, \\
    &E^{c} = \{ (P_{A}(i), P_{B}(j))  |  \forall(i,j) \in D \}.
\end{aligned}
\end{equation}

\subsection{Fine Matches Block (FMB)}
Following LoFTR, we utilize a coarse-to-fine module to derive final fine matches $E^{f} = \{(P_{A}^{f}, P_{B}^{f})\}$.

Specifically, we first locate the position of coarse matches in fine features $\tilde{F}_{A}$, $\tilde{F}_B$ and crop local features $\tilde{F}_{A}^{w}$, $\tilde{F}_{B}^{w} \in \mathbb{R}^{K \times \tilde{C} \times w \times w}$ with a size of $w$.
$K$ means the number of predicted matches.
Then, we reshape the local features to sequences, utilize RecFormer to perform feature aggregation by $L_2$ times and convert the sequences to features.
Subsequently, we calculate the inner product of the transformed features, followed by softmax to generate probability distribution maps.
Finally, we calculate the expectation over the probability distribution as offset $\sigma \in \mathbb{R}^{K \times 2}$ to optimize $P_{B}^{c}$: $P_{B}^{f}$ = $P_{B}^{c}$ + $ \sigma $.
Notably, the matched fine keypoints $P_{A}^{f}$ is the same as $P_{A}^{c}$.

\begin{figure}[]
    \centering
    \includegraphics[width=0.99\hsize]{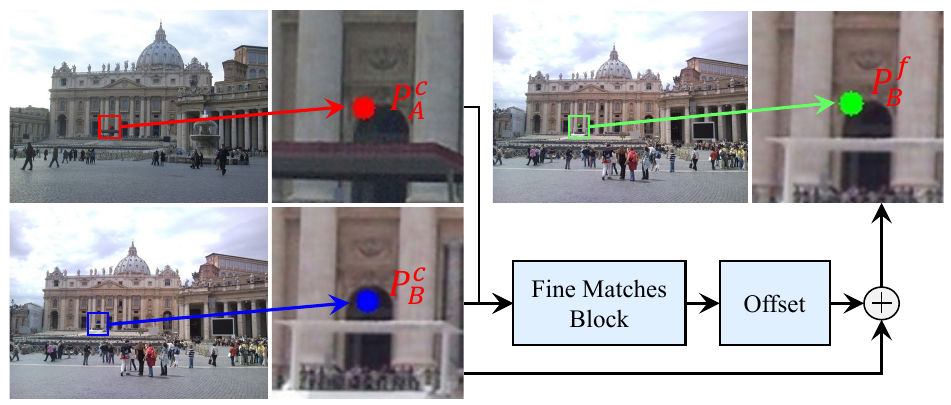}
    \caption{\textbf{The performance of FMB}. Compared with the coarse match $(P_{A}^{c}, P_{B}^{c})$, the fine match $(P_{A}^{f}, P_{B}^{f})$ has less matching errors. Notably, $P_{A}^{f}$ is the same as $P_{A}^{c}$.} 
    \label{fine_matches}
\end{figure}

As shown in \cref{fine_matches}, compared with $P_{B}^{c}$, $P_{B}^{f}$ approximates $P_{A}^{c}$ after being optimized by the predicted offset.

\subsection{Loss}
The total loss $L$ of FMRT consists of two components: (i) Coarse matches loss ${L}_c$ used to supervise the confidence matrix $G$.
(ii) Fine matches loss ${L}_f$ used to supervise the predicted offset $\sigma$.
\begin{equation}
    L = L_{c} + \beta L_{f},
    \label{loss_func}
\end{equation}
where $\beta$ means the weighting coefficient.

\textbf{Definition of Ground-truth Matches $G^{gt}$.}
Following LoFTR, we first compute the index $G^{gt} \in \mathbb{R}$ of the ground-truth matches.
Concretely, we project the keypoints in the left image to the right image and take its nearest neighbor as candidate matches.
The same procedure is employed for the keypoints in the second image.
Finally, we only take the mutual nearest ones as ground-truth matches and define the index of them as $G^{gt}$.

\textbf{Coarse Matches Loss $L_c$.}
We utilize the ground-truth matches $G^{gt}$ to supervise the confidence matrix $G$ and define the coarse matches loss $L_c$ as binary cross entropy:
\begin{equation}
    \begin{aligned}
        L_c&= -[\frac{1}{|G^{gt}|}\sum_{(i,j)\in G^{gt}}\log G(i,j) + \\
        &\frac{1}{N - |G^{gt}|}\sum_{(i,j) \notin G^{gt}}\log (1-G(i,j))].
    \end{aligned}
    \label{4800_loss}
\end{equation}

\textbf{Fine Matches Loss ${L}_f$.}
We first warp the keypoints $P_{A}^{c}$ in the left image to the right image, obtaining $P_{B}^{gt}$. After that, the truth offset $\sigma^{gt} \in \mathbb{R}^{K \times 2}$ is defined as: $\sigma^{gt} = P_{B}^{gt} - P_{B}^{c}$.
Then, we calculate the fine matches loss ${L}_f$ as the L2 norm of $\sigma^{gt}$ and $\sigma$.
\begin{equation}
    \begin{split}
        L_{f} = \frac{1}{K}\sum_{i=1}^{K}\Vert \sigma^{gt}(i)-\sigma(i) \Vert_2,
    \end{split}
\end{equation}
where $K$ denotes the number of predicted matches.

\section{Experiment}

\subsection{Implementation Details}
\textbf{Network Architecture of FMRT.}
For the CNN-based backbone, the feature dimension $\bar{C}$ and $\tilde{C}$ are set to $256$ and $128$, respectively.
For Coarse Matches Block (CMB), the confidence threshold $\rho$ is set to $0.2$.
For Fine Matches Block (FMB), the size $w$ of local windows is set to $5$.
The RecFormer are performed by $L_1=4$ and $L_2=2$ times to enhance features.
When calculating loss, we set the weighting coefficient $\beta$ and predefined threshold $\gamma$ to $0.2$ and $8$, respectively.

\textbf{Training Recipe of FMRT.}
FMRT is trained in an end-to-end manner on the MegaDepth dataset \cite{li2018megadepth} for local feature matching.
The epoch is set to $30$.
Besides, we utilize the AdamW optimizer \cite{loshchilov2017decoupled} with a weight decay of $0.1$.
The learning rate is set to $8\times10^{-3}$, which drops by $0.5$ every $4$ epochs.
We also utilize a linear warmup of $3$ epochs for the learning rate from $8\times10^{-4}$.
Besides, we utilize the gradient clipping technique with a threshold set to $0.5$ to avoid gradient exploding.

\begin{table}[!t]
\centering
\renewcommand\arraystretch{1.2}
\caption{Evaluation on MegaDepth for \textbf{Relative pose estimation}. SP means SuperPoint}
\resizebox{0.49\textwidth}{!}{
\begin{tabular}{clccc}
    \toprule[1pt]
    & \multicolumn{1}{c}{\multirow{2}{*}{Methods}} & \multicolumn{3}{c}{AUC} \\ \cline{3-5} 
    & \multicolumn{1}{c}{} & @5$^{\circ}$     & @10$^{\circ}$     & @20$^{\circ}$   \\ \hline
    \multicolumn{5}{c}{\multirow{1}{*}{Detector-based}}
    \\ \hline                                                                  & SP\cite{DeTone_2018_CVPR_Workshops} + SuperGlue \cite{sarlin2020superglue}\tiny{CVPR'20}                                       & 42.18          & 61.16          & 75.96          \\
    & SP\cite{DeTone_2018_CVPR_Workshops} + ClusterGNN \cite{shi2022clustergnn}\tiny{CVPR'22}                                     & 44.19          & 58.54          & 70.33          \\
    & SP\cite{DeTone_2018_CVPR_Workshops} + DenseGAP \cite{kuang2021densegap}\tiny{ICPR'22}                                     & 41.17          & 56.87          & 70.22          \\
    & SP\cite{DeTone_2018_CVPR_Workshops} + SGMNet \cite{chen2021learning}\tiny{ICCV'21}                                     & 40.50          & 59.00        & 73.60          \\ \hline
    \multicolumn{5}{c}{\multirow{1}{*}{Detector-free}}
    \\ \hline 
    & DRCNet \cite{li2020dual}\tiny{NIPS'20}                  & 27.01  & 42.96  & 58.31          \\ 
    & PDC-Net+ \cite{truong2023pdc}\tiny{TPAMI'23}                                         & 43.10          & 61.90 & 76.10 \\
    & PCFs \cite{zhao2023learning}\tiny{TPAMI'23}                                         & 49.81          & 65.40 & 78.98 \\
    & S2LD \cite{li2023sparse}\tiny{TIP'23}                                      & 49.73 & 65.69 & 78.84         \\
    & LoFTR \cite{sun2021loftr}\tiny{CVPR'21}                                      & 52.80 & 69.19 & 81.18          \\
    & MatchFormer \cite{wang2022matchformer}\tiny{ACCV'22}                                       & 52.91          & 69.74          & 82.00
    \\ 
    & 3DG-STFM \cite{mao20223dg}\tiny{ECCV'22}                                       & 52.58          & 68.46          & 80.04
    \\ 
    & TopicFM \cite{giang2023topicfm}\tiny{AAAI'23}                                       & 54.10          & 70.10          & 81.60
    \\ 
    & QuadTree \cite{tang2022quadtree}\tiny{ICLR'22}                                       & 54.60          & 70.50          & 82.20
    \\ 
    & ASpanFormer \cite{chen2022aspanformer}\tiny{ECCV'22}                                         & 55.30          & 71.50 & 83.10 \\
    & FMRT (ours)              & \textbf{56.42}          & \textbf{72.17} & \textbf{83.54} \\
    \bottomrule[1pt]
\end{tabular}}
\label{outdoor}
\end{table}

\subsection{Relative Pose Estimation}
In this part, we conduct an relative pose estimation experiment on MegaDepth dataset to verify the efficacy of FMRT to predict reliable matches.
Following \cite{sun2021loftr}, we resize the resolution of the images to $840 \times 840$.

Following \cite{sun2021loftr}, we resolution the AUC values of the pose errors under three thresholds $(5^\circ, 10^\circ, 20^\circ)$.

As illustrated in \cref{outdoor}, our proposed FMRT exceeds other cutting-edge methods on relative pose estimation experiment.
Specifically, compared with the detector-based approach ClusterGNN, FMRT boosts the pose estimation precision by $(12.36\%, 13.88\%, 13.40\%)$, which proves the superiority of the detector-free architecture.
Besides, compare with the baseline LoFTR, FMRT achieves an absolute improvement by $(3.75\%, 3.23\%, 2.55\%)$, validating the rationality of adaptively reconciling different receptive fields to generate discriminative features.
Moreover, FMRT surpasses the cutting-edge method QuadTree and ASpanFormer by $(1.25\%, 0.92\%, 0.63\%)$ and $(1.82\%, 1.67\%, 1.34\%)$, further demonstrating the strong feature matching capability of FMRT. 
As shown in \cref{Matcher_Fig}, we exhibit the matching performance comparison between FMRT and LoFTR.
It can be observed that FMRT obtains denser and preciser matches compared to the baseline LoFTR.

\begin{figure}[]
    \centering
    \includegraphics[width=1.0\hsize]{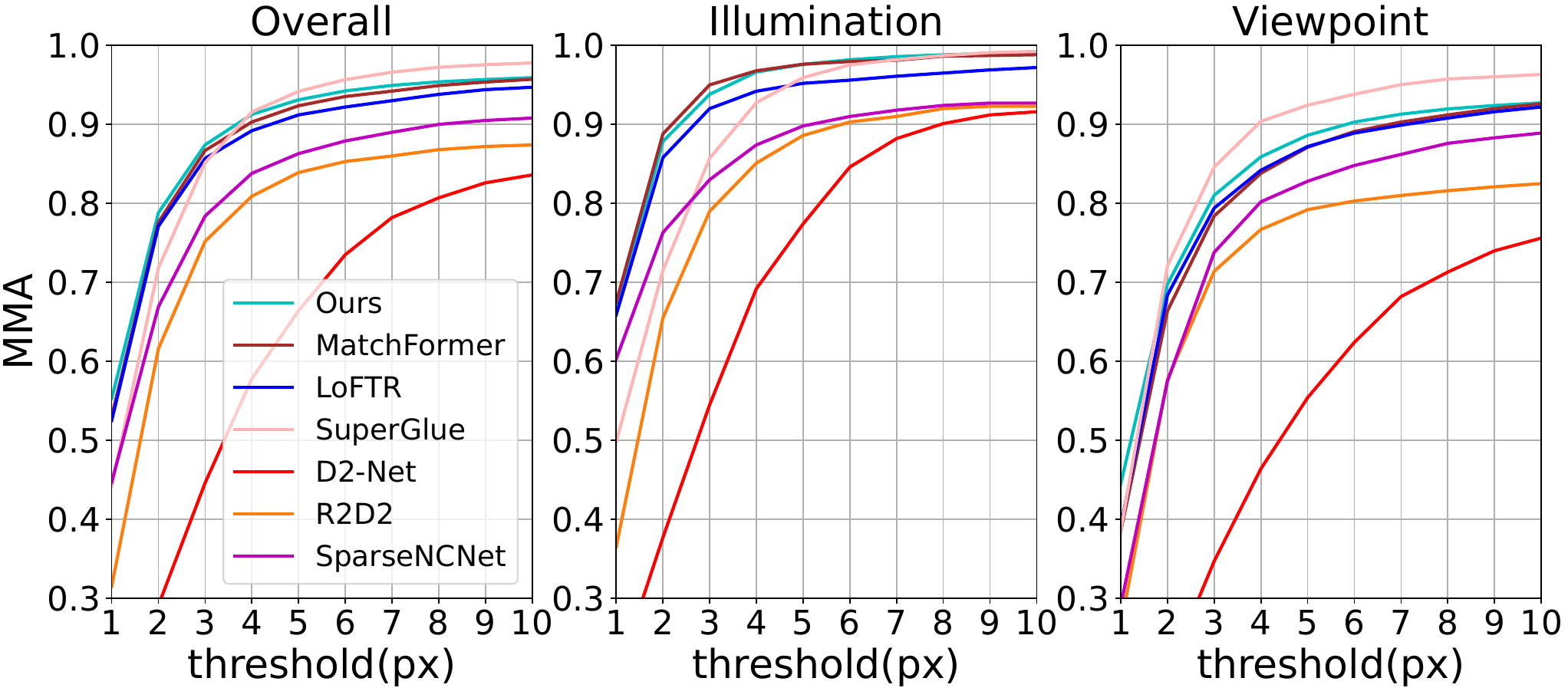}
    \caption{\textbf{Image matching expeiment} on HPatches dataset. The mean matching accuracy (MMA) is reported.} 
    \label{MMA}
\end{figure}
\subsection{Image Matching}
Image matching is a crucial component of substantial computer vision applications.
In this part, we conduct the image matching experiment on the HPatches dataset \cite{balntas2017hpatches} to further exhibit the extraordinary matching capability of FMRT.

Following \cite{Dusmanu_2019_CVPR}, we utilize mean matching accuracy (MMA) as evaluation protocol, which means the proportion of image pairs with reprojection errors is less than a threshold.

\begin{figure}
    \centering
    \includegraphics[width=0.99\hsize]{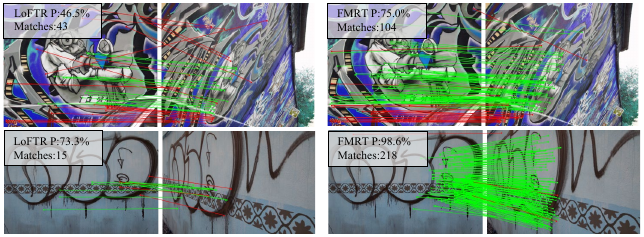}
    \caption{The comparison between LoFTR and FMRT in the HPatches dataset. FMRT exhibits much more robustness under significant viewpoint variations.} 
    \label{Hpatches_Fig}
\end{figure}

Overall, FMRT exhibits impressive matching accuracy compared with the detector-based (i.e., SuperGlue, D2-Net, and R2D2) and detector-free approaches (i.e., SparseNCNet, LoFTR, and MatchFormer).
As illustrated in \cref{MMA}, compared with the baseline LoFTR, FMRT provides competitive performance under extreme illumination conditions.
It is worth noting that FMRT is more robust and reliable under the condition of large viewpoint changes.
As illustrated in \cref{Hpatches_Fig}, FMRT predicts more accurate correspondences compared to the baseline LoFTR.

\begin{table}\Huge
\centering
\renewcommand\arraystretch{1.28}
\caption{Evaluation on HPatches for \textbf{Homography estimation}. The corner correctness metric (CCM) is reported. SP means SuperPoint.}
\resizebox{0.48\textwidth}{!}{
    \begin{tabular}{llccc}
        \toprule[3pt]
        & \multicolumn{1}{c}{\multirow{2}{*}{Methods}} & Overall     & Illumination     & Viewpoint \\ \cline{3-5} 
        & \multicolumn{1}{c}{} & \multicolumn{3}{c}{Accuracy (\%, $\epsilon < 1/3/5$)}   \\ \hline
        \multicolumn{5}{c}{\multirow{1}{*}{Detector-based}}
        \\ \hline
        & D2Net \cite{Dusmanu_2019_CVPR} + NN                            & 0.38/0.71/0.82              & 0.66/0.95/\textbf{0.98} & 0.12/0.49/0.67 \\
        & SP \cite{DeTone_2018_CVPR_Workshops} + NN                                          & 0.46/0.78/0.85              & 0.57/0.92/0.97     & 0.35/0.65/0.74 \\
        & SP \cite{DeTone_2018_CVPR_Workshops} + OANet \cite{zhang2019learning}\Large{CVPR'19}                                          &0.48/0.80/0.86              & 0.58/0.93/0.97 & 0.37/0.68/0.76 \\
        & SP \cite{DeTone_2018_CVPR_Workshops} +SuperGlue \cite{sarlin2020superglue}\Large{CVPR'20}                                   & 0.51/0.82/0.89              & 0.60/0.92/\textbf{0.98} & 0.42/0.71/0.81 \\
        & SP \cite{DeTone_2018_CVPR_Workshops} + ClusterGNN \cite{shi2022clustergnn}\Large{CVPR'22}                                         & 0.52/\textbf{0.84}/0.90              & 0.61/0.93/\textbf{0.98} & \textbf{0.44/0.74}/0.81 \\ 
        \hline
        \multicolumn{5}{c}{\multirow{1}{*}{Detector-free}}
        \\ \hline     
        & Patch2Pix \cite{zhou2021patch2pix}\Large{CVPR'21}                                   & 0.50/0.79/0.87              & 0.71/0.95/\textbf{0.98} & 0.30/0.64/0.76 \\
        & SparseNCNet \cite{rocco2020efficient}\Large{ECCV'20}                                 & 0.36/0.65/0.76              & 0.62/0.92/0.97 & 0.13/0.40/0.58 \\
        & LoFTR \cite{sun2021loftr}\Large{CVPR'21}                                         & \textbf{0.55}/0.81/0.86              & 0.74/0.95/\textbf{0.98} & 0.38/0.69/0.76 \\
        & ASpanFormer \cite{chen2022aspanformer}\Large{ECCV'22}                                & 0.54/0.82/0.90              & 0.70/0.95/0.98 & 0.38/0.70/0.81 \\
        & MatchFormer \cite{wang2022matchformer}\Large{ACCV'22}                                & \textbf{0.55}/0.81/0.87              & \textbf{0.75}/0.95/\textbf{0.98} & 0.37/0.68/0.78 \\
        & TopicFM \cite{giang2023topicfm}\Large{AAAI'23}                                & 0.55/0.82/0.89              &  0.74/\textbf{0.96/0.98} & 0.37/0.68/0.78 \\
        & FMRT (ours)                                       & \textbf{0.56}/0.82/\textbf{0.91}           & 0.73/0.95/\textbf{0.98}  & 0.40/0.71/\textbf{0.83} \\ \bottomrule[3pt]
\end{tabular}}
\label{homo_graphy}
\end{table}

\subsection{Homography Estimation}
In this part, we appraise FMRT in the homography estimation on the HPatches benchmark.
Following \cite{zhou2021patch2pix}, we utilize corner correctness metric (CCM) as metrics, which means the percentage of images whose corner errors are less than thresholds.

In \cref{homo_graphy}, FMRT realizes the extraordinary performance among all methods overall under the threshold of $1$ and $5$ pixels.
More concretely, FMRT exceeds the MatchFormer and baseline LoFTR with the boost of $(1\%, 1\%, 4\%)$ and $(1\%, 1\%, 5\%)$.
This shows that reconciling features with different receptive fields is conducive to generating discriminative features used to predict accurate matches.
Furthermore, FMRT surpasses detector-free methods Patch2Pix, LoFTR, and MatchFormer by $(10\%, 7\%, 7\%)$, $(2\%, 2\%, 7\%)$, and $(3\%, 3\%, 5\%)$ under extreme viewpoint changes, which further validates the robustness of FMRT.

\begin{table}[] \Large
\centering
\renewcommand\arraystretch{1.2}
\caption{Evaluation on InLoc for \textbf{Visual localization evaluation}. SP means SuperPoint.}
\resizebox{0.46\textwidth}{!}{
    \begin{tabular}{lccc}
        \toprule[2.0pt]
        \multicolumn{1}{c}{}                         & DUC1                                      & DUC2                                      \\ \cline{2-3} 
        \multicolumn{1}{c}{\multirow{-2}{*}{Method}} & \multicolumn{2}{c}{(0.25m,10)/(0.5m,10)/(1.0m,10)}                                    & \multicolumn{1}{c}{\multirow{-2}{*}{AP}} \\ \hline
        \multicolumn{3}{c}{Detector-based Methods}                                                                                           \\ \hline
        D2Net \cite{Dusmanu_2019_CVPR}+NN               &  38.4/56.1/71.2                        & 37.4/55.0/64.9       & 53.8                \\
        SP \cite{detone2018superpoint}+NN               &  38.4/56.1/71.2                        & 37.4/55.0/64.9       & 53.8                \\
        SP \cite{DeTone_2018_CVPR_Workshops}+SuperGlue \cite{sarlin2020superglue}\small{CVPR'20}        & 49.0/68.7/80.8                        & 53.4/\textbf{77.1}/82.4     & 68.6                   \\
        SP \cite{DeTone_2018_CVPR_Workshops}+SGMNet \cite{chen2021learning}\small{ICCV'21}        & 41.9/64.1/73.7                        & 39.7/62.6/67.2         & 58.2               \\
        SP \cite{DeTone_2018_CVPR_Workshops}+ClusterGNN \cite{shi2022clustergnn}\small{CVPR'22}       & 47.5/69.7/79.8                        & 53.4/\textbf{77.1}/84.7        & 68.7               \\
        SP \cite{DeTone_2018_CVPR_Workshops}+LightGlue \cite{shi2022clustergnn}\small{ICCV'23}       & 49.0/68.2/79.3                        & 55.0/74.8/79.4        & 67.6               \\ \hline
        \multicolumn{3}{c}{Detector-free Methods}                                                                                            \\ \hline
        Patch2Pix \cite{zhou2021patch2pix}\small{CVPR'21}                                     & 44.4/66.7/78.3                        & 49.6/64.9/72.5     & 62.7                   \\
        LoFTR \cite{sun2021loftr}\small{CVPR'21}                                       & 47.5/72.2/84.8                        & 54.2/74.8/\textbf{85.5}     & 69.8                   \\
        MatchFormer \cite{wang2022matchformer}\small{ACCV'22}                                 & 46.5/73.2/\textbf{85.9}                        & \textbf{55.7}/71.8/81.7        & 69.1                \\
        S2LD \cite{li2023sparse}\small{TIP'23}                                 & 46.5/73.2/\textbf{85.9}                        & \textbf{55.7}/71.8/81.7        & 69.1                \\
        FMRT (ours)                        & \textbf{52.0/73.7}/83.3   & 54.2/73.3/84.0
        & \textbf{70.0} \\ \bottomrule[2.0pt]
\end{tabular}}
\label{inloc_table}
\end{table}

\subsection{Visual Localization}
Feature matching is an essential part of visual localization \cite{taira2018inloc, xie2022deep, dai2023eaainet, sarlin2019coarse}.
In this part, we integrate our network into an official visual localization pipeline HLoc \cite{sarlin2019coarse} and conduct an experiment on the InLoc dataset \cite{taira2018inloc}.
Notably, we utilize FMRT trained on the MegaDepth and submit the predicted pose to the Long-Term Visual Localization Benchmark \cite{toft2020long} to derive the accuracy.

Follow \cite{sun2021loftr, wang2022matchformer}, we utilize the DUC1 and DUC2 that report the percentage of images with localization errors less than thresholds as metrics.
Besides, we calculate the average of DUC1 and DUC2 as average precision (AP).

As illustrated in \cref{inloc_table}, FMRT realizes the best localization precision in terms of DUC1 $(0.25m,10)$, $(0.5m,10)$, and AP values.
Specifically, FMRT outperforms the detector-based methods SuperGlue, SGMNet, and ClusterGNN in almost all metrics and boosts the AP values by $1.4\%$, $11.8\%$, and $1.3\%$.
Besides, FMRT outstrips the baseline MatchFormer and LoFTR by $0.9\%$ and $0.2\%$ in terms of AP values, further proving its superior matching capability.

\begin{figure*}[]
    \centering
    \includegraphics[width=1.0\hsize]{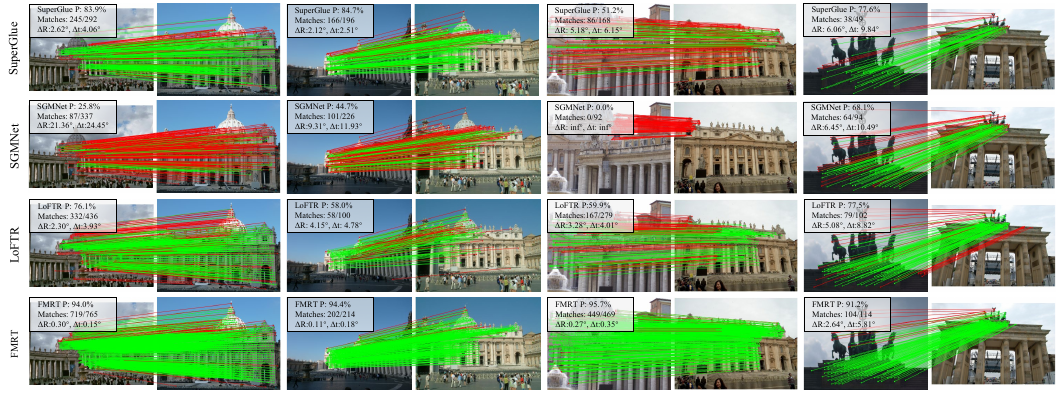}
    \caption{Qualitative matching under among SuperGlue, SGMNet, LoFTR, and FMRT.
    FMRT shows extraordinary matching ability under challenging conditions such as extreme illumination and viewpoint variation.} 
    \label{Matcher_Fig}
\end{figure*}

\subsection{Understanding FMRT}

\begin{table}[!t]
	\centering
	\renewcommand\arraystretch{1.2}
	\caption{\textbf{Efficiency analysis.} We report runtime (s) and AUC values of several detector-free methods.}
	\resizebox{0.44\textwidth}{!}{
		\begin{tabular}{l|ccc}
        \toprule[1.0pt]
        \multicolumn{1}{c|}{Methods}  & Runtime (s) & AUC@(5$^{\circ}$, 10$^{\circ}$, 20$^{\circ}$)  \\ \midrule
        Patch2Pix \cite{zhou2021patch2pix}    & 0.684 & 41.40 / 56.30 / 68.32      \\
        LoFTR \cite{sun2021loftr}                        & \textbf{0.176} & 52.80 / 69.19 / 81.18      \\
        QuadTree \cite{tang2022quadtree}                 & 0.296    & 54.60 / 70.50 / 82.20  \\
        MatchFormer \cite{wang2022matchformer}                    & 0.826      & 52.91 / 69.74 / 82.00  \\
        FMRT                      & 0.260  &  \textbf{56.42} / \textbf{72.17} / \textbf{83.54}       \\
        \bottomrule[1.0pt]
		\end{tabular}
	}
	\label{time_flops}
\end{table}
\textbf{Runtime Evaluation.}
We evaluate the runtime of the proposed FMRT and compare it with other cutting-edge detector-free methods.
We randomly select $100$ image pairs from the MegaDepth dataset and resize the images to $1024 \times 1024$ for Patch2Pix, $840 \times 840$ for LoFTR, QuadTree, MatchFormer, and FMRT.
As illustrated in \cref{time_flops}, LoFTR has a speedy inference speed at the cost of inferior localization performance.
Besides, compared with other detector-free methods (i.e., QuadTree and MatchFormer), FMRT achieves extraordinary performance with $(12.16\%, 68.52\%)$ inference speed boost.

\begin{figure}[]
    \centering
    \includegraphics[width=1.0\hsize]{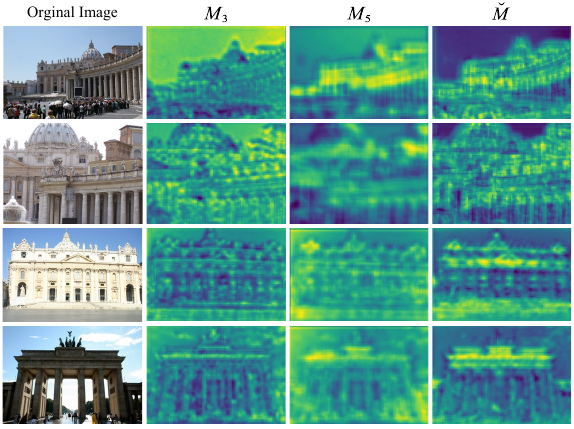}
    \caption{Visualization of the features $M_{3}$, $M_{5}$ and the reconciled features $\breve{M}$. The reconciled features $\breve{M}$ contain abundant geometry curves and semantic information.} 
    \label{reconcile_feature}
\end{figure}
\textbf{Visualization of the features $M_{3}$, $M_{5}$, and $\breve{M}$.}
To investigate the rationality of reconciling features with multiple receptive fields, we visualize the features $M_{3}$, $M_{5}$, and $\breve{M}$.
As shown in \cref{reconcile_feature}, the features $M_{3}$ and $M_{5}$ mainly contain abundant geometry curves and semantic information, respectively.
FMRT utilizes the Perception Weight Layer (PWL) to ensure the network automatically weighs the importance of features with different receptive fields.
As a consequence, the reconciled features $\breve{M}$ incorporate both structure curve and semantic information.

\textbf{Qualitative Matching Results Comparison.}
As shown in \cref{Matcher_Fig}, we exhibit a qualitative comparison among SuperGlue, SGMNet, FMRT and the baseline LoFTR.
We can observe that FMRT noticeably outperforms the baseline LoFTR with more precise and dense matches in challenging conditions (e.g., extreme illumination and viewpoint change).

\begin{figure*}
    \centering
    \includegraphics[width=1.0\hsize]{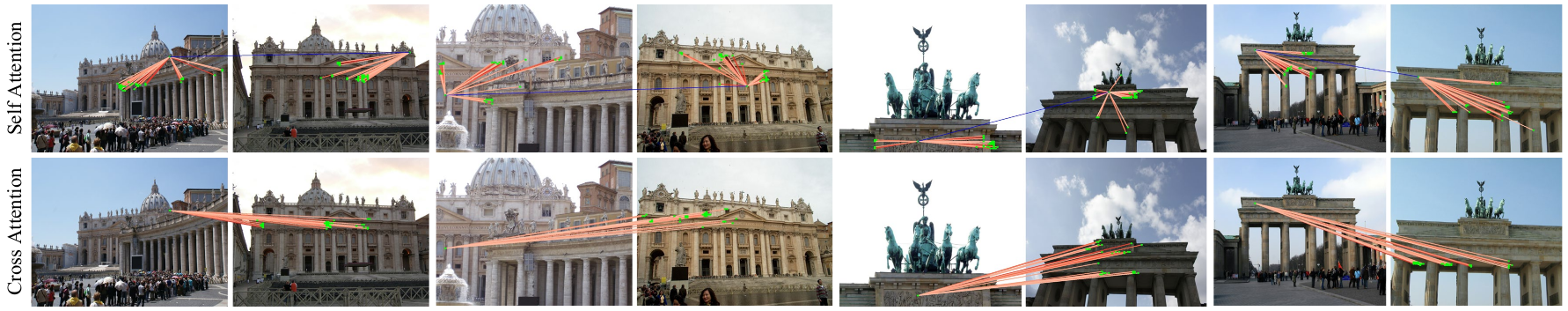}
    \caption{Visualizing self-attention and cross-attention. Four image pairs are selected from the MegaDepth dataset. the query point mainly aggregates the global context information from the corners and edges.}
    \label{attention_line_fig}
\end{figure*}
\textbf{Attention Weights Analysis.}
In this part, we show the attention weight in \cref{attention_line_fig}.
Specifically, we select a pair of keypoints and calculate the attention matrix $\phi(Q)\phi(K)^{T}$ to select $30$ keypoints with the largest response values.
We can observe that each keypoint focuses on the edges of objects to aggregate the surrounding global context information.

\begin{table}[]\Huge
    \centering
    \renewcommand\arraystretch{1.12}
    \caption{Ablation study for \textbf{different proposed modules}.}
    \resizebox{0.46\textwidth}{!}{
        \begin{tabular}{l|ccc}
            \toprule[2.5pt]
            \multicolumn{1}{c|}{Methods}          & AUC@5$^\circ$ & AUC@10$^\circ$ & AUC@20$^\circ$ \\ \midrule
            w/o AWPE         & 55.48      & 71.47       & 82.96       \\
            Using Linear Attention \cite{katharopoulos2020transformers}         & 54.61      &  70.86      & 82.17       \\
            w/o PWL         & 55.90      & 71.89       & 83.14       \\
            w/o DW convolution in MLFFN        & 55.27      & 71.29       & 82.41       \\
            w/o FMB          & 52.87      & 69.13       & 80.87       \\
            FMRT full &  \textbf{56.42}  &\textbf{72.17}  &\textbf{83.54}        \\ \bottomrule[2.5pt]
        \end{tabular}}
    \label{ablation_conponment}
\end{table}

\begin{table}[]\large
    \centering
    \renewcommand\arraystretch{1.20}
    \caption{Ablation study for \textbf{different positional encoding methods}.}
    \resizebox{0.46\textwidth}{!}{
        \begin{tabular}{l|ccc}
            \toprule[1.5pt]
            \multicolumn{1}{c|}{Methods}          & AUC@5$^\circ$ & AUC@10$^\circ$ & AUC@20$^\circ$ \\ \midrule
            Positional Encoding in \cite{sun2021loftr}         & 56.08   & 71.75    &  83.28     \\
            Positional Encoding in \cite{sarlin2020superglue}         & 55.95   & 71.86    &  83.36     \\
            Axis-Wise Position Encoder     & \textbf{56.42}      & \textbf{72.17}       & \textbf{83.54}       \\
            \bottomrule[1.5pt]
        \end{tabular}}
    \label{ablation_positional}
\end{table}

\subsection{Ablation Study.}
To verify the effectiveness of different components of FMRT, we conduct an ablation experiment for different variants of FMRT on the MegaDepth dataset. 

\textbf{The effect of different modules.}
As shown in \cref{ablation_conponment}, we can observe that all of the proposed components elevate the pose estimation accuracy.
(i) Discarding Axis-Wise Position Encoder (AWPE) leads to $(0.94\%, 0.70\%, 0.58\%)$ accuracy reduction.
(ii) Replacing the Global Perception Attention Layer (GPAL) and Perception Weight Layer (PWL) with the linear attention \cite{katharopoulos2020transformers} makes the accuracy drop by $(1.81\%, 1.31\%, 1.37\%)$, demonstrating that using features with different receptive fields to integrate global context information is beneficial.
(iii) Removing the Perception Weight Layer (PWL) spawns a drop $(0.52\%, 0.28\%, 0.40\%)$. proving that reconciling multiple receptive fields adaptively is conducive to learning discriminative features.
(iv) Discarding the depth-wise convolution in Local Perception Feed-forward Network (LPFFN) results in $(1.15\%, 0.88\%, 1.13\%)$ accuracy reduction.
(v) Removing Fine Matches Block (FMB) leads to a significantly lower accuracy $(3.55\%, 3.04\%, 2.67\%)$, proving the effectiveness of optimizing coarse matches.

\textbf{The effect of different positional encoding methods.}
To verify the proposed Axis-Wise Position Encoder (AWPE), we compared it with other positional encoding methods proposed in \cite{sun2021loftr} and \cite{sarlin2020superglue}.
More concretely, we separately integrate different encoding methods into FMRT and conduct outdoor pose estimation experiments on the MegaDepth dataset.
As shown in \cref{ablation_positional}, compared with the positional encoding used in LoFTR, the proposed Axis-Wise Position Encoder boosts the matching performance by $(0.34\%, 0.32\%, 0.26\%)$, proving the superiority of using network to extract reliable positional encoding information.
Moreover, AWPE achieves better performance than the positional encoding used in SuperGlue, demonstrating the effectiveness of viewing the positional encoding as two independent keypoints encoding tasks along the row and column dimensions.

\begin{table}[]\large
    \centering
    \renewcommand\arraystretch{1.20}
    \caption{Ablation study for \textbf{different combinations of depth-wise convolutions}.}
    \resizebox{0.40\textwidth}{!}{
        \begin{tabular}{l|ccc}
            \toprule[1.5pt]
            \multicolumn{1}{c|}{Methods}          & AUC@5$^\circ$ & AUC@10$^\circ$ & AUC@20$^\circ$ \\ \midrule
            $(3\times3, 7\times7)$         & 56.13   & 71.82    &  83.33     \\
             $(5\times5, 7\times7)$      & 55.83   & 71.77    &  83.08     \\
          $(3\times3, 5\times5)$    & \textbf{56.42}      & \textbf{72.17}       & \textbf{83.54}       \\
            \bottomrule[1.5pt]
        \end{tabular}}
    \label{ablation_dw}
\end{table}

\textbf{The effect for different combinations of depth-wise convolutions.}
To verify the rationality of the combination of depth-wise convolutions employed in the Feature Perception Layer, we utilize other combinations (i.e., $(3\times3, 7\times7)$, $(5\times5, 7\times7)$) to present the relative pose estimation experiment.
As illustrated in \cref{ablation_dw}, applying the combination of $(3\times3, 5\times5)$ brings superior performance.

\begin{table}[]\large
    \centering
    \renewcommand\arraystretch{1.20}
    \caption{Ablation study for \textbf{different weighting coefficient $\beta$}.}
    \resizebox{0.35\textwidth}{!}{
        \begin{tabular}{l|ccc}
            \toprule[1.5pt]
            \multicolumn{1}{c|}{Methods}          & AUC@5$^\circ$ & AUC@10$^\circ$ & AUC@20$^\circ$ \\ \midrule
            $\beta=1$          & 52.92   & 69.75    &  81.83     \\
             $\beta=0.5$      & 54.82   & 70.43    &  83.06     \\
          $\beta=0.2$ & \textbf{56.42}      & \textbf{72.17}       & \textbf{83.54}       \\
          $\beta=0.1$ & 56.06      & 71.63       & 83.36       \\
            \bottomrule[1.5pt]
        \end{tabular}}
    \label{ablation_loss}
\end{table}
\textbf{The effect of different weighting coefficient $\beta$.}
As shown in \cref{loss_func}, we utilize weighting coefficient $\beta$ In this part, we perform an ablation experiment with different  weighting factor $\beta$.
As illustrated in \cref{ablation_loss}, FMRT achieves the best localization precision when $\beta$ is set to $0.2$.
When $\beta$ is too large, the model has a bias target towards the refinement process, which makes the confidence matrix weakly supervised.
When $\beta$ is too small, FMRT cannot effectively optimize the predicted coarse matches, resulting in inferior localization performance.

\section{Conclusion}
In this study, we propose a novel detector-free method FMRT that extracts reliable and precise correspondences.
FMRT proposes a dedicated Reconciliatory Transformer that ensures the network adaptively measures the importance of features with different receptive fields, hence boosting the image perception capability of the network.
Besides, FMRT introduces a straightforward Axis-Wise Position Encoder (AWPE) that perceives the positional encoding as two independent keypoints encoding tasks along the row and column dimensions and utilizes a neural network to realize excellent position encoding.
Comprehensive experiments demonstrate that FMRT achieves extraordinary performance in several tasks, such as relative pose estimation, homography estimation, image matching, and visual localization.

\bibliographystyle{IEEEtran}
\bibliography{IEEEabrv,bib.bib}

\end{document}